%% file: main.tex
\documentclass[10pt,twocolumn,letterpaper]{article}

\usepackage[pagenumbers]{wacv}


\definecolor{wacvblue}{rgb}{0.21,0.49,0.74}
\usepackage[pagebackref,breaklinks,colorlinks,allcolors=wacvblue]{hyperref}
\usepackage{multirow}
\usepackage{multicol}
\usepackage[table]{xcolor}
\usepackage[dvipsnames]{xcolor}
\usepackage{tabularx}

\title{Judging by Appearances? Auditing and Intervening\\ Vision-Language Models for Bail Prediction}

\author{
    Sagnik Basu\textsuperscript{\rm 1} \qquad
    Shubham Prakash\textsuperscript{\rm 1} \qquad
    Ashish Maruti Barge\textsuperscript{\rm 1} \qquad
    Siddharth D Jaiswal\textsuperscript{\rm 1} \\
    Abhisek Dash\textsuperscript{\rm 2} \qquad
    Saptarshi Ghosh\textsuperscript{\rm 1} \qquad
    Animesh Mukherjee\textsuperscript{\rm 1} \\
    \textsuperscript{\rm 1}Indian Institute of Technology Kharagpur, India \\
    \textsuperscript{\rm 2}Max Planck Institute for Software Systems, Saarbruecken, Germany \\
    \tt\small \textsuperscript{\rm 1}\{basusagnik99.24, shubham.0313, ashishbarge.24, siddsjaiswal\}@kgpian.iitkgp.ac.in \\
    \tt\small \textsuperscript{\rm 1}\{animeshm, saptarshi\}@cse.iitkgp.ac.in \\
    \tt\small \textsuperscript{\rm 2}adash@mpi-sws.org
}

\begin{document}
\maketitle
\input{sec/0_abstract}
\input{sec/1_intro}
\input{sec/2_rwork}
\input{sec/3_models_data}
\input{sec/4_exps}
\input{sec/5_results}
\input{sec/6_discn}
\input{sec/7_end}
{
    \small
    \bibliographystyle{ieeenat_fullname}
    \bibliography{main}
}

\end{document}

%% file: sec/0_abstract.tex
\begin{abstract}
Large language models (LLMs) have been extensively used for legal judgment prediction tasks based on case reports and crime history. However, with a surge in the availability of large vision language models (VLMs), legal judgment prediction systems can now be made to leverage the images of the criminals in addition to the textual case reports/crime history. Applications built in this way could lead to inadvertent consequences and be used with malicious intent. In this work, we run an audit to investigate the efficiency of standalone VLMs in the bail decision prediction task. We observe that the performance is poor across multiple intersectional groups and models \textit{wrongly deny bail to deserving individuals with very high confidence}. We design different intervention algorithms by first including legal precedents through a RAG pipeline and then fine-tuning the VLMs using innovative schemes. We demonstrate that these interventions substantially improve the performance of bail prediction. Our work paves the way for the design of smarter interventions on VLMs in the future, before they can be deployed for real-world legal judgment prediction.
\end{abstract}

%% file: sec/1_intro.tex
\section{Introduction}
\label{sec:intro}
Artificial Intelligence (AI) is showing significant promise for transforming the legal landscape, with research exploring its application in automating tasks like legal document summarization~\cite{akter2025comprehensive}, legal information retrieval~\cite{sansone2022legal}, legal judgment prediction (LJP)~\cite{malik2021ildc} and many more. 
Among the most consequential ones is LJP, where machine learning or deep learning models predict judicial outcomes ---such as bail decisions~\cite{morin2024machine}, conviction status, or sentencing~\cite{lyu2023multi,malik2021ildc}--- by reasoning over case facts, statutes (a formal, written law) and precedents~\cite{wu2023precedent} (relevant previous cases). 
LJP has been extensively studied across several jurisdictions, with applications ranging from Supreme People's Court of China~\cite{yuan2019automatic} to U.S. Supreme Court decisions~\cite{katz2017general}, as well as emerging applications in India~\cite{malik2021ildc} and Australia~\cite{shareghi2024methods}.
In recent times, the complex reasoning process required for LJP has been transformed by the advent of large language models (LLMs) which enable richer analyses of these legal documents~\cite{jiang2023legal}.

If we consider an LJP scenario in a real courtroom setup (where a case is being argued in a court of Law), there are multiple modalities involved with the decision making process, such as images and videos as evidence of the crime (\eg~face photographs, evidence from CCTV footage), speech data from the witnesses, as well as the textual data (\eg~FIRs, judgments, witness statements in text). However, existing studies in LJP rely heavily on textual data, such as court opinions, statutes and case briefings for modelling judicial outcomes. The days are not far when vision language models (VLMs)~\cite{qwen2.5-VL,liu2024llavanext,wang2025internvl3_5,laurençon2024building}, which can understand and integrate information from both image and text domains, would very likely be used to enhance AI-assisted legal decision making.
In fact, recently, VLMs have been used to scan and process legal documents. For instance, they have been used to understand first information report (FIR) forms, which are often hand-written and scanned~\cite{chakraborty2025well,westermann2024analyzing}. Another area where VLMs are being extensively used is in forensic image analysis, where it has been observed that such AI tools cannot replace forensic experts~\cite{Farber2025ForensicAI} and can at best work as an assistive tool.

While VLMs are being increasingly used for various tasks in the legal and other domains, they have certain limitations as well.
\textit{Bias issues} in VLMs are an active area of concern~\cite{ruggeri2023multi,xiao2024genderbias}, especially as these models are increasingly applied in sensitive domains like healthcare~\cite{hartsock2024vision}, education~\cite{stamatakis2025enhancing} and law~\cite{chakraborty2025well}. Contini \etal~\cite{Contini2024AI_Legal} discuss how the emergence of such generative AI tools has reduced the delivery of justice to mere statistical correlations among data, thus completely suspending the emotive-cognitive process that lies at the core of legal decision making. These issues raise critical questions about fairness, accountability and generalizability of AI systems in legal applications.

One of the specific tasks in AI assisted LJP is \textit{bail decision prediction}, wherein the model predicts whether an accused person can be granted bail or not based on multiple facts and evidences.
This is a binary prediction task, where the output 0/no indicates that the accused should be denied bail, and 1/yes indicates that the accused should be granted bail.
As in a real-world scenario, multiple modalities are involved, to further reduce workloads in courts, VLMs have much potential to be used in these setups where a face image is provided along with the case texts. 
This work tries to ``foresee'' such a future scenario and evaluate the VLMs on the basis of their known pitfalls. In this paper, we particularly focus on three core research questions:\\
\noindent
\textbf{RQ1. How does a VLM behave in the presence of different modalities (\ie image and text) in legal bail prediction?}
This research question focuses on different VLMs and their performance when it is presented with one or both of the image and text modalities.\\
\textbf{RQ2. Does incorporating previous case reports mirroring the common law system help improve the models' ability to make accurate and consistent decisions?} Many countries adopt the common law system, in which judges issue their judgments
based on statutes as well as precedents. To address this question, we devise a RAG-based framework that acts as a database containing previous case reports, and VLMs base their predictions on the retrieved relevant cases.\\
\textbf{RQ3. How do the RAG setup and fine-tuning of VLMs interact to influence the bail decision prediction?} Since fine-tuning is known to improve the prediction performance of VLMs, we investigate whether vanilla or sophisticated fine-tuning coupled with the already developed RAG can improve the bail decision prediction performance.\\
\textbf{Key contributions}: The key contributions of our paper are as follows.
\begin{enumerate}
    \item We pair two datasets - (i) a dataset of mugshots of accused persons and (ii) a dataset of case reports with corresponding bail decisions to create input instances to audit the performance of VLMs in legal judgment prediction.
    \item We audit the standalone VLMs to test their performance in predicting bail decisions when the input is a pair consisting of an image of an accused person and the associated case report.
    \item We perform several interventions on the VLMs, which are described in \cref{sec:exps}. We show that by applying the correct interventions, we can bring out better performance from a VLM in a legal judgment prediction context. These interventions primarily include RAG based precedent inclusion and different fine-tuning schemes for the VLMs.
    \item Aside from the accuracy metric, we also evaluate the chosen VLMs based on LR- and NPV, as these metrics measure the likelihood of false negative predictions and the trustworthiness of the models, which highly matters in a legal context since it is considered more important to ensure that an innocent individual is not denied bail, than to ensure that a criminal is not granted bail. From our comprehensive audit of multiple VLMs, we show that all the models in their base form perform very poorly in accuracy as well as in the LR- and NPV metrics. The most alarming part is that, on average, \textit{these models are highly confident in} $\sim \mathbf{68\%}$ \textit{of their false negative predictions}. Nevertheless, suitably designed intervention techniques result in \textit{steady and substantial gains} in terms of all three metrics. This leads us to believe that if VLMs are indeed to be used for LJP in future, effective intervention techniques can be built before they are deployed for this task.
\end{enumerate}

%% file: sec/2_rwork.tex
\section{Related work}
\label{sec:relwork}

In this section, we review the existing literature regarding the innovations and contributions in LJP over the past few years. We also show how the use of many AI tools and techniques facilitate the legal process and law enforcement. At the end of this section we also discuss recent advancements in VLMs and their potential for future applications in legal domain.

\subsection{Text-focused LJP}
The field of LJP has evolved from early machine learning models using TF-IDF on case facts~\cite{aletras2016predicting} to more sophisticated deep learning approaches. A significant leap occurred with the adoption of the transformer architectures, particularly with the development of domain-specific pre-trained language models like Legal-BERT~\cite{chalkidis2020legal} and InLegalBERT~\cite{paul-2022-pretraining}
which demonstrate superior performance by pre-training on large legal corpora. 
The current state-of-the-art is driven by large language models (LLMs), with research shifting toward fine-tuning both general-purpose models~\cite{yuan2019automatic,jiang2023legal} and specialized legal LLMs like ChatLaw~\cite{cui2023chatlaw} to handle the nuanced reasoning required for legal tasks.
Nigam \etal~\cite{nigam2024rethinking} show through their evaluation on Llama2 (70B and 13B)~\cite{touvron2023llama} and GPT3.5-Turbo~\cite{brown2020language} that prompting the LLMs with the facts as well as statutes, precedents, rulings by lower courts and arguments to mimic real-world scenarios significantly enhances the quality of the predictions.

\subsection{Dataset contributions in LJP}
Progress in LJP has been heavily dependent on the creation of high-quality datasets. Foundational benchmarks include the European Court of Human Rights (ECHR) corpus~\cite{poudyal2020echr} and the large-scale Chinese criminal case dataset, CAIL~\cite{xiao2018cail2018}. For the Indian legal system, the Indian Legal Documents Corpus (ILDC)~\cite{malik2021ildc} and the Hindi Legal Documents Corpus (HLDC)~\cite{kapoor2022hldc} are two of the most crucial resources that provides a benchmark for understanding the specifics of Indian case law, which we also leverage to build our customised dataset. Our contribution extends this by creating a novel evaluation setup that pairs legal text with the Illinois DOC labeled faces dataset, enabling the study of multimodal LJP in a way that existing text-only legal datasets do not support. A very recent work by Nigam \etal~\cite{nigam2024nyayaanumana} proposes the largest and most diverse dataset of Indian legal cases, alongside a specialised language model fine-tuned for LJP and explainability. A very contemporary work on NyayaRAG~\cite{nigam2025nyayarag} builds the foundation by showing that structured legal retrieval enhances both outcome accuracy and interpretability.

\subsection{Recent applications of VLMs}
VLMs have matured from establishing shared image-text embeddings with CLIP~\cite{radford2021learning} to full-fledged conversational agents like LLaVA~\cite{liu2023visual}, which connect powerful vision encoders to LLMs. The advanced models we employ -- LLaVA-NeXT~\cite{liu2024llavanext}, Qwen2.5-VL~\cite{qwen2.5-VL}, Idefics3~\cite{laurençon2024building}, and InternVL 3.5~\cite{wang2025internvl3_5} -- represent this frontier, leveraging strong LLM backbones like Mistral~\cite{jiang2023mistral7b} and Llama3~\cite{dubey2024llama}. 
While these models have seen rapid adoption in high-stakes domains such as medicine for analyzing radiological images~\cite{hartsock2024vision}, their application in the socio-technical and high-stakes legal domain has only begun recently~\cite{chakraborty2025well,westermann2024analyzing}. 
Our research is among the first to systematically audit these modern VLMs in the context of legal judgment prediction.

%% file: sec/3_models_data.tex
\section{Models \& Datasets}

\begin{table*}[t!]
\centering
\begin{tabularx}{\textwidth}{>{\centering\arraybackslash}m{0.2\textwidth} X}
\textbf{Image} & \textbf{Case facts} \\
\hline

\includegraphics[width=0.4\linewidth]{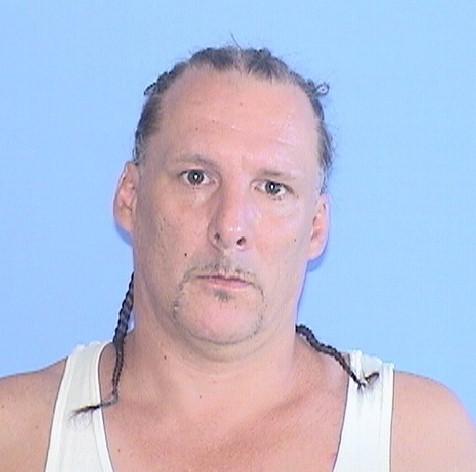} &
...The accused has no prior criminal history. $< Name >$ has no independent $< Name >$...the accused was carrying the animal in the said truck and when his truck was checked, \textbf{two bundles of two kilograms of charas} were recovered from it... \\

\includegraphics[width=0.4\linewidth]{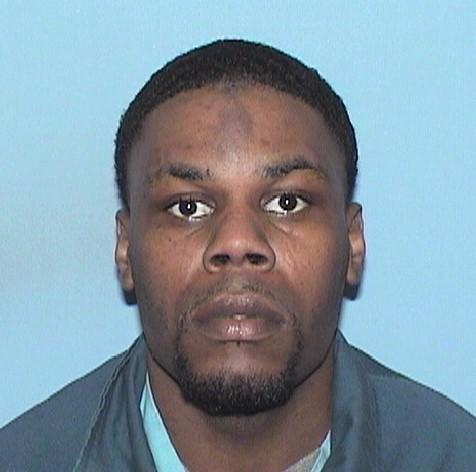} &
...Despite the occurrence of the incident, there are no independent public witnesses. They have been in jail since the date of 29-12-2019. Around 130 kilograms of beef, bovine remains head, hoof, legs, peacock skin and cow-\textbf{slaughter tools were recovered from the spot}.... \\

\includegraphics[width=0.4\linewidth]{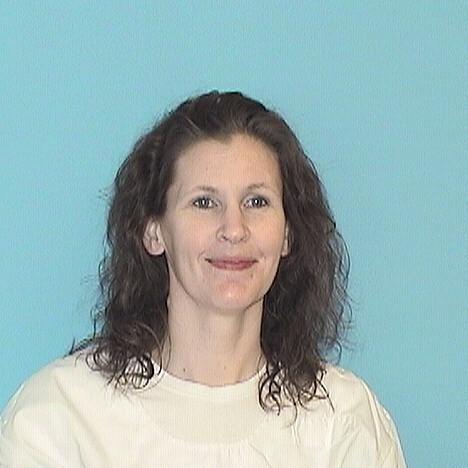} &
...and not do any work in the name of marriage and used to talk for hours on the phone $< name >$ to which the applicant / accused objected and threatened that these days girls are heard a lot, \textbf{you will be framed in a false case and put in jail, false case filed}... \\

\includegraphics[width=0.4\linewidth]{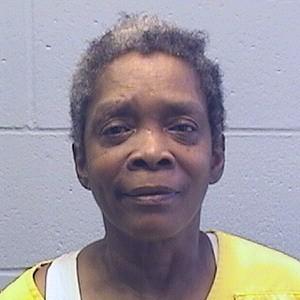} &
...There is \textbf{no recovery of any illegal asset from the possession of the candidate} which is alleged to have been recovered. It is false and fabricated... \\
\hline
\end{tabularx}
\caption{Combination of Illinois Images paired with case facts. Due to limited space, only a relevant portion of the case facts is shown.}
\label{fig:img_paired_text}
\end{table*}

Our experiments are built upon a curated set of two distinct datasets, each serving a specific role, and four state-of-the-art VLMs in the 7-8B parameter range, selected for their advanced capabilities in understanding both image and text modalities.

\subsection{Models}
\label{sec:models}
We select four powerful, open-source, instruction-tuned VLMs that represent the current state-of-the-art.
\begin{itemize}
    \item \textbf{LLaVA-NeXT}~\cite{liu2024llavanext}: We use the 7B parameter variant, specifically \texttt{llava-v1.6-mistral-7b-hf}.
    It pairs a strong CLIP-based vision encoder with the \textbf{Mistral-7B-Instruct-v0.2} LLM backbone.

    \item \textbf{Qwen2.5-VL}~\cite{qwen2.5-VL}: We employ the \texttt{Qwen2.5-VL-7B-Instruct} model that is paired with the \textbf{Qwen2.5-7B-Instruct} LLM.

    \item \textbf{Idefics3}~\cite{laurençon2024building}: We use the \texttt{Idefics3-8B-Llama3} model
    that is built upon the \textbf{Llama3.1-8B-Instruct} LLM backbone, paired with the SigLip vision model.
    
    \item \textbf{InternVL3.5}~\cite{wang2025internvl3_5}: We use the \texttt{InternVL3\_5-8B-HF} model that follows a ``ViT-MLP-LLM'' paradigm, pairing a powerful InternViT vision encoder with the LLM from the \textbf{Qwen3} series.
\end{itemize}

\subsection{Datasets}
\label{sec:datasets}

\begin{itemize}
    \item \textbf{Illinois DOC labeled faces dataset~\cite{illinoisimages}:} 
    contains prisoner mugshots from the Illinois Department of Corrections. Crucially for our audit, it provides associated metadata for each image, including sensitive Personally Identifiable Information (PII) such as \textbf{race} label (\eg ``Black'', ``White'', ``Hispanic'', ``Asian'' \etc) and \textbf{gender} label (``male'', ``female'').
    For this work, we have only chosen images from four intersectional groups -- ``White Male'' (WM), ``Black Male'' (BM), ``White Female'' (WF) and ``Black Female'' (BF) as Whites and African Americans are the majority among other races.
    In this dataset, the male-female ratio among the ``Whites'' is $13370 : 1628$ and among the ``Blacks'' is $28156 : 1240$. The dataset also has information about \textbf{offense types} associated with the accused, which can be grouped into the following broad categories -- \textit{weapons violation, theft, battery, narcotics, homicide, burglary, robbery, motor vehicle theft, intimidation, stalking, criminal trespass, liquor law violation, prostitution, human trafficking, public indecency, assault, public peace violation}.

    \item \textbf{Legal documents corpus~\cite{kapoor2022hldc}:} To obtain the case reports, we utilize the development set of the HLDC~\cite{kapoor2022hldc}, containing 17K legal case documents. This corpus is based on the \texttt{Exploration-Lab/IL-TUR} benchmark. The dataset contains case reports with the corresponding ground-truth bail decisions in Hindi. Note that this is the only available comprehensive dataset of bail decisions and thus for our purpose we translate the dataset into English using IndicTrans2~\cite{gala2023indictrans2} which is a popular translation model between English and Indian languages. We select only the \texttt{dev\_all} split (with 17,707 case reports) from the entire dataset for our use case. Following Kapoor et al. \cite{kapoor2022hldc} we consider only the \texttt{facts\_and\_arguments} column of the dataset as the case report. Next, we perform the following preprocessing steps.
    \begin{enumerate}
    \item We prompt GPT-4o~\cite{achiam2023gpt} to give us a list of common keywords found in criminal case reports, such as ``suspect'', ``case number'', ``arrest''~\etc. Following the stopword removal procedure in NLP, we remove such words from the dataset as they act as stopwords in a legal context.
    \item We tokenize each ``facts and arguments'' using \texttt{Mistral-7B-Instruct-v0.2}~\cite{jiang2023mistral7b} and remove any cases that have a token length less than 50.
    \item From the \texttt{facts\_and\_arguments} column we extract only the facts as having the arguments in the case reports can introduce bias in the input to the VLM. To this purpose, we select some keywords, such as ``oppose'', ``granted'', ``rejected''~\etc and remove the sentences containing them. We check if the token length becomes less than 50 and remove the corresponding case reports from the dataset.
\end{enumerate}
After this preprocessing step, the dataset size becomes 16,104. We split the data further into train (12,788 cases) and test (3,316 cases) set in $80:20$ ratio. Note that in our RAG framework, the training samples serve as the external knowledge base. Henceforth, we shall use the term \textit{case fact} instead of case report as we extract only the facts from the full case reports and use it for all our experiments.
\end{itemize}

\subsection{Pairing criminal images with case reports}
For all our experiments, we pair the image dataset with the case fact dataset since there is no benchmark dataset available with such paired information. In particular, we pair each case fact with all the images in each intersectional group. Thus, if there are $N$ images in all and $M$ case facts, then the total number of pairs we have is $N\times M$. Finally if the case fact is from the training dataset then all the pairs in which it participates are included into the training data; similarly, if the case fact is from the test dataset then all the pairs in which it participates become part of the test data. \cref{fig:img_paired_text} shows some examples of image-case fact pairs from our dataset.

%% file: sec/4_exps.tex
\section{Experimental setup}
\label{sec:exps}
This section details the experimental innovations of our work based on the three research questions.
First we present the framework for auditing VLMs in the presence of different modalities, such as image and text (\textbf{RQ1}). Next, to address the second research question (\textbf{RQ2}) we evaluate the models with a context-aware RAG setup. Finally, we describe the comprehensive pipeline of various fine-tuning schemes for the bail prediction task (\textbf{RQ3}).

\subsection{Auditing VLMs for legal judgment prediction}
\label{sec:vlm_audit}
We choose a VLM, $\mathcal{M}$, to audit its prediction performance across the intersectional group comprising gender and race. Let the paired image ($I$) and case fact ($C_{TST}$) be denoted as $[I:C_{TST}]$. As noted earlier, the test set comprises all pairs in which an instance of $C_{TST}$ is a part of. We then pass this pair as input to $\mathcal{M}$ and query it for the bail decision as follows.  
\centerline{$\mathcal{M}([I:C_{TST}]) = \textrm{yes} | \textrm{no}$}
where yes (no) indicates that $\mathcal{M}$ predicts that bail should (should not) be granted. These predictions are then compared with the ground truth to compute various metrics discussed later in this section.

\subsection{Intervention I: Precedent-aware VLMs}
\label{sec:rag_vlm}
In many countries including India, the legal system follows the \textit{Common Law} paradigm~\cite{fullerton2006evolution}, which is grounded in precedent and judicial decisions rather than solely codified statutes. In this paradigm, the judges interpreting past cases that were similar to the current case, and apply the judgements/principles in those prior cases to the new/current case. Motivated by this, we investigate \textit{whether incorporating similar past case facts as precedents improves model predictions and, in turn, enhances fairness in bail decisions} across different intersectional groups.

To operationalize the common law paradigm in our setting, we design a RAG framework~\cite{zhao2024retrieval,asai2023self,wang2024speculative}. Specifically, we construct a vector store from the training set ($C_{TRN}$) of the case facts, and
make the model first retrieve relevant precedents and then generate responses taking into account the retrieved similar precedents.
Using a Euclidean distance–based similarity measure, we retrieve the top three relevant case facts closest to the query case fact and append them to the system prompt, along with the query case fact in the user prompt, to obtain bail decision predictions from the model.
This can be expressed as:
\centerline{$\mathcal{M}([RAG(C_{TRN}) : I : C_{TST}]) = \textrm{yes} | \textrm{no}$}

\begin{figure}[t]
    \centering
    \includegraphics[width=0.95\linewidth, height=0.5\linewidth]{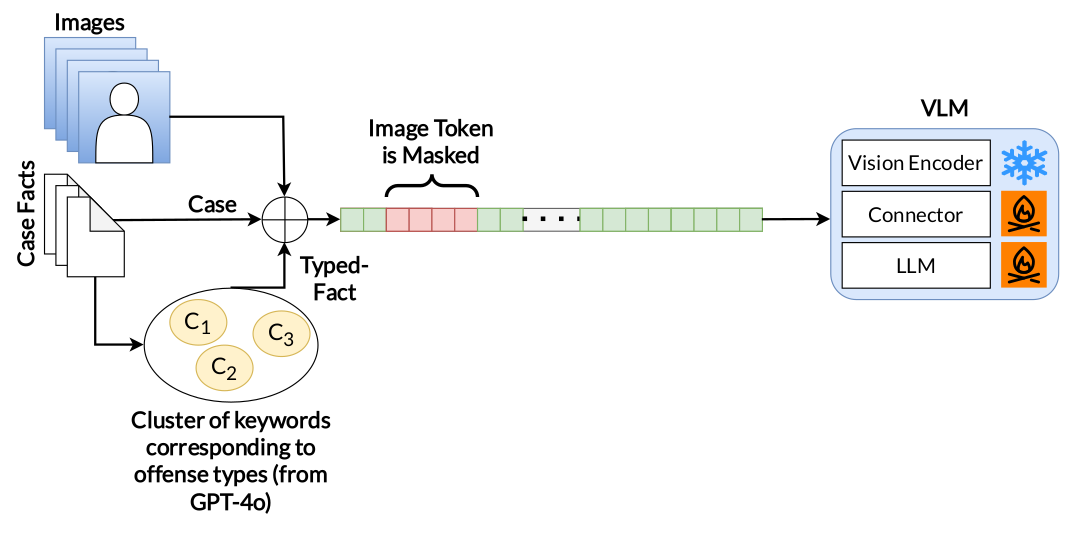}
    \caption{Overall fine-tuning design. For \textit{vanilla} setup typed-facts are not added; however, in \textit{offense type induced} setup, the whole architecture is in use.}
    \label{fig:finetune_flow}
\end{figure}

\subsection{Intervention II: Fine-tuned VLMs}
\label{sec:finetune_vlm}
To adapt VLMs for the bail prediction task, we develop a comprehensive supervised fine-tuning (SFT) pipeline. The primary motivation behind this step is to bridge the gap between the generic pre-training objectives of VLMs and the highly domain-specific nature of bail decision prediction task. 
The fine-tuning process begins with input construction, which we design in two ways: (1) \textit{vanilla}: using only the case facts as the user prompt, and (2) \textit{offense type induced}: to begin with, we utilise the offense types included in the metadata of the Illinois dataset as discussed in \cref{sec:datasets}. For each offense type, we obtain similar keywords by querying GPT-4o~\cite{achiam2023gpt}. Thus, for example, the offense type \textit{homicide} expands to the set of following semantically similar keywords: \{\textit{homicide, murder, manslaughter, first-degree murder, second-degree murder, $\dotsc$}\}. Similarly, the offense type \textit{theft} expands to \{\textit{theft, grand theft, shoplifting, burglary, pickpocketing, $\dotsc$}\} while the offense type \textit{narcotics} expands to \{\textit{controlled substance, cocaine, heroin, methamphetamine, marijuana, $\dotsc$}\}. Now we consider each case fact and search for the presence of keywords from each of these expanded sets. If a case fact is found to contain one or more keywords corresponding to an offense type, then that offense type is associated with the case fact. Note that more than one offense type might get associated with a case fact in this process. We call such a type associated case fact as \textit{typed-fact}. In this way, all the case facts in all the training and test pairs are converted to typed-facts. These typed-facts are used to form the user prompt. \cref{fig:finetune_flow} shows the full fine-tuning setup. In the \textit{vanilla} design, we do not add the typed-facts to the user prompt. In the \textit{offense type induced} design, only the typed-facts are included in the user prompt following the cases, as shown in \cref{fig:finetune_flow}.

In both setups, like earlier, the user prompt is paired with an image $I$ from the Illinois dataset as visual evidence. For loss computation, we frame the task as a binary decision-making problem with a supervised textual target (yes/no), which is the prediction output.
During fine-tuning we make the parameters of the vision tower frozen,
and completely mask the images from the input by setting the values in the attention mask corresponding to the image tokens (\ie \texttt{<image>}) to 0. 
Through this approach, we want to ensure that \textit{the models learn how the case facts lead to bail acceptance or rejection, not which person is linked with which case fact}. Since the image attentions are 0, for a given case fact/typed-fact, while the training data contains $N$ images with which it is paired up, we use one random pair out of these, as the case fact only matters in the fine-tuning while the pairing image does not.

Fine-tuning is performed with the \texttt{SFTTrainer} from the TRL library, using the \texttt{adamw\_torch} optimizer, a learning rate of $1\times 10^{-5} $, and an effective batch size of 8 through gradient accumulation. Model selection is guided by validation loss minimization, where the validation set is taken to be 10\% of the train set described in \cref{sec:datasets}. We do a full fine-tuning on the models, \ie updating all parameters except the vision parameters, maximizing task-specific adaptation. Let the vanilla fine-tuned model be denoted as $\mathcal{M}^{V}$ and the one based on offense type be denoted as $\mathcal{M}^{O}$. 
The predictions on the test set are then obtained using the two different models in the following ways.
\begin{itemize}
    \item $\mathcal{M}^{V}([I : C_{TST}]) = \textrm{yes} | \textrm{no}$
    \item $\mathcal{M}^{V}([RAG(C_{TRN}): I : C_{TST}]) = \textrm{yes} | \textrm{no}$
    \item $\mathcal{M}^{O}([RAG(C_{TRN}^{TYPE}): I : C_{TST}^{TYPE}]) = \textrm{yes} | \textrm{no}$
\end{itemize}
Here $C_{TRN}^{TYPE}$ and $C_{TST}^{TYPE}$ respectively correspond to the typed-facts in the train and test sets.
\subsection{Evaluation metrics}
\label{sec:metrics}

We evaluate the VLMs for their performance on the bail prediction task across three metrics defined below. For each of the metrics, the positive prediction is ``bail granted'' and the negative prediction is ``bail denied''.

\noindent
$\bullet$ \textbf{Accuracy.}
This metric is calculated as $\frac{TP + TN}{TP + FP + TN + FN}$ and captures a high-level perspective of how well the VLMs perform for the task of bail prediction aggregated over all the intersectional groups.\\
\noindent
$\bullet$ \textbf{Negative likelihood ratio ($\mathbf{LR-}$).} This is calculated as $\frac{FNR}{TNR}$, where $FNR = \frac{FN}{TP + FN}$ and $TNR = \frac{TN}{TN + FP}$. Here, $LR-$ represents the likelihood of bail being denied to individuals who are, in fact, eligible for release as per the ground-truth.\\
\noindent
$\bullet$ \textbf{Negative predictive value (NPV).}
Defined as $\frac{TN}{TN + FN}$, it measures the probability of bail denials being correct, thereby indicating the trustworthiness of bail denial decisions.

The choice of the metrics $LR-$ and NPV are motivated by the following argument --  in tasks like granting of bail in the legal domain, one should favor a decision making process in which the false negative rate is low~\cite{zafar:17}, \ie, it is more important to ensure that a person deserving bail is not denied of it than to ensure that a criminal is not granted bail\footnote{``\textit{It is better that ten guilty persons escape than that one innocent
suffer}'' -- William Blackstone.}.

%% file: sec/5_results.tex
\section{Results}
\label{sec:res}
In this section, we discuss the results from our experiments.
Table~\cref{tab:all_results} shows the results (described below) where each metric is reported individually for the four intersectional groups -- ``White Male'' (WM), ``Black Male'' (BM), ``White Female'' (WF) and ``Black Female'' (BF).

\begin{table*}
    \centering
    \small
    \begin{tabular}{c|c|c|c|c|ccc}
        \toprule
        \multirow{2}{*}{\textbf{Models}} & \multicolumn{2}{c|}{\multirow{2}{*}{\textbf{Metrics}}} & \multicolumn{1}{c|}{\textbf{Audit}} & \multicolumn{1}{c|} {\textbf{Intervention I}} & \multicolumn{3}{c}{\textbf{Intervention II: Fine-tuning}}\\\cmidrule{4-8}
         & & & $\mathcal{M}$ & $\mathcal{M}[RAG]$ & $\mathcal{M}^{V}$ & $\mathcal{M}^{V}[RAG]$ & $\mathcal{M}^{O}[RAG]$ \\
        \midrule
        \multirow{9}{*}{Qwen} & \multicolumn{2}{c|}{Overall accuracy ($\uparrow$)} & 41.96\% & \cellcolor{OliveGreen!60}{58.10\%} & 40.62\% & 65.92\% & \cellcolor{OliveGreen!60}{68.03}\% \\
        \cmidrule{2-8}
        & \multirow{4}{*}{LR- ($\downarrow$)} & WM & 0.97 & \cellcolor{Cerulean!75}0.68 & 0.96 & 0.53 & \cellcolor{Cerulean!75}0.47 \\
        & & BM & 0.97 & \cellcolor{Cerulean!75}0.69 & 0.95 & 0.55 & \cellcolor{Cerulean!75}0.48 \\
        & & WF & 0.96 & \cellcolor{Cerulean!75}0.68 & 0.96 & 0.52 & \cellcolor{Cerulean!75}0.49 \\
        & & BF & 0.97 & \cellcolor{Cerulean!75}0.68 & 0.95 & 0.52 & \cellcolor{Cerulean!75}0.48 \\
        \cmidrule{2-8}
        & \multirow{4}{*}{NPV ($\uparrow$)} & WM & 38.01\% & \cellcolor{GreenYellow!75}46.70\% & 38.21\% & 52.53\% & \cellcolor{GreenYellow!75}55.59\% \\
        & & BM & 38.20\% & \cellcolor{GreenYellow!75}46.37\% & 38.64\% & 52.34\% & \cellcolor{GreenYellow!75}55.39\% \\
        & & WF & 38.23\% & \cellcolor{GreenYellow!75}46.86\% & 38.49\% & 53.34\% & \cellcolor{GreenYellow!75}54.96\% \\
        & & BF & 38.10\% & \cellcolor{GreenYellow!75}46.81\% & 38.33\% & 53.02\% & \cellcolor{GreenYellow!75}55.31\% \\
        \midrule
        \multirow{9}{*}{Llava-NeXT} & \multicolumn{2}{c|}{Overall accuracy ($\uparrow$)} & 48.08\% & \cellcolor{OliveGreen!60}{52.40\%} & 72.44\% & 74.27\% & \cellcolor{OliveGreen!60}{75.72\%} \\
        \cmidrule{2-8}
        & \multirow{4}{*}{LR- ($\downarrow$)} & WM & 0.89 & \cellcolor{Cerulean!75}0.78 & 0.44 & 0.35 & \cellcolor{Cerulean!75}0.27 \\
        & & BM & 0.89 & \cellcolor{Cerulean!75}0.79 & 0.44 & 0.35 & \cellcolor{Cerulean!75}0.27 \\
        & & WF & 0.87 & \cellcolor{Cerulean!75}0.76 & 0.4 & 0.34 & \cellcolor{Cerulean!75}0.26 \\
        & & BF & 0.86 & \cellcolor{Cerulean!75}0.76 & 0.4 & 0.33 & \cellcolor{Cerulean!75}0.26 \\
        \cmidrule{2-8}
        & \multirow{4}{*}{NPV ($\uparrow$)} & WM & 40.12\% & \cellcolor{GreenYellow!75}43.20\% & 57.45\% & 62.72\% & \cellcolor{GreenYellow!75}68.49\% \\
        & & BM & 40.13\% & \cellcolor{GreenYellow!75}43.12\% & 57.85\% & 63.29\% & \cellcolor{GreenYellow!75}68.75\% \\
        & & WF & 40.60\% & \cellcolor{GreenYellow!75}44.00\% & 59.97\% & 63.46\% & \cellcolor{GreenYellow!75}69.73\% \\
        & & BF & 41.01\% & \cellcolor{GreenYellow!75}43.88\% & 59.80\% & 63.85\% & \cellcolor{GreenYellow!75}69.20\% \\
        \midrule
        \multirow{9}{*}{Idefics3} & \multicolumn{2}{c|}{Overall accuracy ($\uparrow$)} & 46.79\% & \cellcolor{OliveGreen!60}{56.96\%} & 56.78\% & 69.70\% & \cellcolor{OliveGreen!60}{72.74\%} \\
        \cmidrule{2-8}
        & \multirow{4}{*}{LR- ($\downarrow$)} & WM & 0.9 & \cellcolor{Cerulean!75}0.71 & 0.73 & 0.46 & \cellcolor{Cerulean!75}0.36 \\
        & & BM & 0.92 & \cellcolor{Cerulean!75}0.72 & 0.72 & 0.45 & \cellcolor{Cerulean!75}0.35 \\
        & & WF & 0.89 & \cellcolor{Cerulean!75}0.72 & 0.68 & 0.46 & \cellcolor{Cerulean!75}0.36 \\
        & & BF & 0.9 & \cellcolor{Cerulean!75}0.73 & 0.7 & 0.44 &\cellcolor{Cerulean!75} 0.32 \\
        \cmidrule{2-8}
        & \multirow{4}{*}{NPV ($\uparrow$)} & WM & 39.55\% & \cellcolor{GreenYellow!75}45.32\% & 44.87\% & 56.39\% & \cellcolor{GreenYellow!75}62.01\% \\
        & & BM & 39.34\% & \cellcolor{GreenYellow!75}45.51\% & 45.41\% & 56.88\% & \cellcolor{GreenYellow!75}63.20\% \\
        & & WF & 40.20\% & \cellcolor{GreenYellow!75}45.45\% & 46.89\% & 56.31\% & \cellcolor{GreenYellow!75}62.75\% \\
        & & BF & 39.55\% & \cellcolor{GreenYellow!75}44.79\% & 45.69\% & 56.99\% & \cellcolor{GreenYellow!75}64.76\% \\
        \midrule
        \multirow{9}{*}{InternVL} & \multicolumn{2}{c|}{Overall accuracy ($\uparrow$)} & 58.10\% & \cellcolor{OliveGreen!60}{65.26\%} & 65.88\% & 68.18\% & \cellcolor{OliveGreen!60}{68.52\%} \\
        \cmidrule{2-8}
        & \multirow{4}{*}{LR- ($\downarrow$)} & WM & 0.75 & 0.54 & \cellcolor{Cerulean!75}0.42 & 0.47 & 0.46 \\
        & & BM & 0.72 & 0.54 & \cellcolor{Cerulean!75}0.39 & 0.48 & 0.46 \\
        & & WF & 0.7 & 0.54 & \cellcolor{Cerulean!75}0.35 & 0.44 & 0.43 \\
        & & BF & 0.71 & 0.52 & \cellcolor{Cerulean!75}0.39 & 0.44 & 0.43 \\
        \cmidrule{2-8}
        & \multirow{4}{*}{NPV ($\uparrow$)} & WM & 43.89\% & \cellcolor{GreenYellow!75}52.18\% & \cellcolor{GreenYellow!75}58.61\% & 55.61\% & 56.47\% \\
        & & BM & 45.39\% & \cellcolor{GreenYellow!75}52.62\% & \cellcolor{GreenYellow!75}60.37\% & 55.70\% & 56.45\% \\
        & & WF & 45.84\% & \cellcolor{GreenYellow!75}52.62\% & \cellcolor{GreenYellow!75}63.32\% & 57.47\% & 58.10\% \\
        & & BF & 45.27\% & \cellcolor{GreenYellow!75}53.04\% & \cellcolor{GreenYellow!75}60.23\% & 57.35\% & 57.85\% \\
        \bottomrule
    \end{tabular}
    \caption{Results of the experiments. Column 4 presents the results of the audit of the VLMs. Column 5 presents the results of Intervention I (precedent-aware VLMs), and columns 6-8 present the results of Intervention II (various types of fine-tuning).
     The best values for each of the three metrics are color-coded as follows: \colorbox{OliveGreen!60}{\textbf{Overall accuracy}}, \colorbox{Cerulean!75}{\textbf{Group-based LR-}} and \colorbox{GreenYellow!75}{\textbf{Group-based NPV}} (best viewed in color). $\downarrow$: lower is better, $\uparrow$: higher is better.}
    \label{tab:all_results}
\end{table*}

\subsection{Results of auditing standalone VLMs (RQ1)}
We present the results of auditing the standalone VLMs in Table~\cref{tab:all_results}. The fourth column, labelled as $\mathcal{M}$, notes the accuracy, LR- and NPV values for this audit. For all the VLMs, barring InternVL, the accuracy is below 50\%. The LR- values are very high for all the intersectional groups, and the NPV values are no better than 45\%. This means that irrespective of the intersectional group, \textit{a large majority of deserving individuals (as per the ground truth) are denied bail}. 
We further ask the VLM about its confidence (high/medium/low) in the judgment predictions. On average across the models, out of all cases where a bail is incorrectly denied, in as many as $\sim 68\%$ cases, the VLM does it with high confidence. This is a severely alarming trend, making the base VLMs unsuitable for LJP.

\subsection{Results of Intervention I (RQ2)}
The results of the precedent-aware VLM are presented in the fifth column of the \cref{tab:all_results} labelled as $\mathcal{M}[RAG]$. For all the VLMs, we observe a steady improvement in the accuracy, while there is a remarkable increase of 16.14\% for Qwen. Overall, the best accuracy of 65.26\% is obtained for InternVL. More importantly, for all the models and across all intersectional groups, the $LR-$ values have declined, and the NPV values have increased. Thus, Intervention~I has made the VLMs much more suitable for the bail prediction task. 

\subsection{Results of Intervention II (RQ3)} 
We evaluate the fine-tuned VLMs in three setups -- (a) $\mathcal{M}^{V}$, \ie using only the fine-tuned VLM without any precedents (column 6, \cref{tab:all_results}), (b) $\mathcal{M}^{V}[RAG]$, retrieving relevant case reports using a RAG and adding them into the user prompt while querying the fine-tuned model (column 7, \cref{tab:all_results}), and (c) $\mathcal{M}^{O}[RAG]$, retrieving relevant typed-facts using the RAG and adding them in the user prompt while querying the model that is also fine-tuned with typed-facts (column 8, \cref{tab:all_results}). In all cases, $\mathcal{M}^{O}[RAG]$ by far outperforms all the other intervention mechanisms in terms of accuracy. For Llava-NeXT and Idefics3, the accuracy values cross the 70\% mark, with Llava-NeXT reporting as high as 75.72\% accuracy. In terms of LR-, there is a drastic reduction in the case of $\mathcal{M}^{O}[RAG]$ compared to the other intervention schemes for Qwen, Llava-NeXT and Idefics3. For InternVL, $\mathcal{M}^{V}$ has the least LR- while $\mathcal{M}^{O}[RAG]$ is close second. In terms of NPV, once again $\mathcal{M}^{O}[RAG]$ achieves the best values compared to the other intervention schemes for Qwen, Llava-NeXT and Idefics3. For InternVL, $\mathcal{M}^{V}$ has the best NPV while $\mathcal{M}^{O}[RAG]$ is again close second.
Evidently, both our interventions are effective in making VLMs more appropriate for the bail prediction task.

\vspace{3mm}
\noindent A general observation for the base models as well as for the intervention schemes is that males experience slightly higher false negative outputs than females, \ie $LR-$ values for females are marginally lower than those of males, whereas the NPV values for females are marginally higher than those for males. In other words, for all the setups, \textit{males are denied bail slightly more often than females}.

%% file: sec/6_discn.tex
\section{Discussion}
In this work, we perform a comprehensive \textit{anticipatory audit} of VLMs for the task of bail prediction (one of the critical tasks in the stack of LJP tasks). We prepare a large-scale multimodal dataset of case facts~\cite{kapoor2022hldc} and mugshot images of suspects~\cite{illinoisimages} (across four intersectional groups), which are then used to evaluate the VLMs under multiple settings. 

\subsection{Qualitative observations}
The initial audit, on vanilla out-of-the-box VLMs reveals that the models perform very poorly, with accuracy as low as 42\%. The fairness metrics, while not having any statistically significant differences between the intersectional groups, report poor absolute values. We conclude that not only are the models more likely to reject bail to those deserving of it (high LR-), but also that these bail decisions are not trustworthy (low NPV). Finally, a disturbing trend is that in around 68\% of cases, these models are highly confident in rejecting bail to deserving candidates. Thus, using such models without domain knowledge baked into the pipeline is highly dangerous in sensitive contexts like legal AI.

\begin{table*}[t]
\centering
\tiny
\begin{tabular}{|p{0.07\textwidth}|p{0.2\textwidth}|p{0.68\textwidth}|}
\hline
\textbf{Models} & \textbf{Current case (excerpts)} & \textbf{Precedent cases retrieved (excerpts)} \\
\hline
\multicolumn{3}{|c|}{\textbf{\texttt{no} to \texttt{yes} corrections}} \\
\hline
$\mathcal{M}^O[RAG]$ & ... the accused objected and threatened that these days girls are heard a lot, you will be framed in a false case and put in jail, false case filed ... 
& .... FIR and medical are contradictory. The applicant / accused has no previous criminal history ...	\newline
... accused has been falsely implicated in the said false case by a conspiracy only to harass and humiliate him / her ...	\newline
... the plaintiff had borrowed two thousand rupees from the applicant / accused on a weekly basis. The plaintiff did not return the money and was pressurized ...	\\
\hline
 $\mathcal{M}^V[RAG]$ & ... It is not possible to kidnap any girl student from the school from the said place ...
&  ... to have occurred at about 12 noon... $<$name$>$ was alone with her father ... came to the house in the name of $<$name$>$ and started to protest ... \newline There is an old enmity between the applicant and the plaintiff ... That is why a false case of the applicant / accused is written under pressure ...\\

\hline
\hline
\multicolumn{3}{|c|}{\textbf{\texttt{yes} to \texttt{no} corrections}} \\
\hline
$\mathcal{M}^O[RAG]$ & ... when his truck was checked, two bundles of two kilograms of charas were recovered from it ... 
&  ... does not have a criminal history. The applicant has been shown to have made a false recovery... \newline ... a plastic foil was removed from the right side of the shirt pants ... The accused is stated to have recovered 500 grams of charas ... \newline
... $<$name$>$ selling ganja to the persons going to come in front of his house. When $<$name$>$ was stopped, he behaved indecently with the policemen ...
\\
\hline
$\mathcal{M}^V[RAG]$ & ... the accused is alleged to have killed $< name >$ ... and thrown his body on the railway line of Jungle Gram Kaili ...
& ... We brought the injured $< name >$ to the road $< name >$ and from there went to the hospital ... Dr $< name >$ saw that $< name >$ was dead ... \newline
... The family members slept in their house at night ... uncle of the plaintiff, gave information from his phone number to the plaintiff's brother that someone had killed ... \newline
... which he started looking for and met his friend $< name >$ and $< name >$ told him that his son had left home and had not yet reached home ...
\\
\hline
\end{tabular}
\caption{Examples of current cases and retrieved precedent cases, grouped by whether predictions correctly changed after an intervention from rejection to granting of bail or from granting to rejection. Only small excerpts shown since case documents are very long.}
\label{tab:precedent_cases}
\end{table*}

Next, we introduce two simple, but highly useful and relevant, interventions into the pipeline -- (i) using a precedent-aware VLM where the precedents are brought in from a vector store of relevant case records and (ii) fine-tuned VLMs under different fine-tuning schemes.
We note an immediate improvement in all metrics for both interventions. First, using only the precedents on a vanilla VLM not only improves the accuracy by as much as 16\%, but also has an impact on the fairness metrics. Interestingly, the improvement in fairness metrics is, again, (almost) independent of the intersectional groups. Finally, the second intervention, which involves a supervised fine-tuning of the VLM shows a marked improvement on all metrics when the precedent-aware RAG is incorporated into the pipeline. 
In \cref{tab:precedent_cases}, we note some of the representative errors that get corrected for each of the interventions. The 1$^\textrm{st}$ and 3$^\textrm{rd}$ rows show the cases rectified after using $\mathcal{M}^O[RAG]$ and 2$^\textrm{nd}$ and 4$^\textrm{th}$ rows show rectifications after using $\mathcal{M}^V[RAG]$. It is evident from these examples that the retrieved cases are relevant, which influences the predictions for the current case. We also note that the retrieved cases after using $\mathcal{M}^O[RAG]$ are very accurate and aligned with the offense type of the current case. Such enhanced retrievals help the VLMs reach the correct judgment, which is otherwise not possible for a base model.

Finally, we caution here that while the models, upon intervention, perform better compared to their standalone versions, the absolute accuracies are still at best 76\%, and further work is needed before they can be deployed in the real world for sensitive legal AI tasks.

\subsection{Position: Possibility of using VLMs for bail prediction}
Legal AI is a rapidly growing field with adoption rates increasing throughout the world~\cite{yuan2019automatic,katz2017general,malik2021ildc,shareghi2024methods}. Commonly, AI tools are being used in the courtrooms to aid the legal system through transcription~\cite{mahapatra2023milpac}, legal statute identification~\cite{paul2024legal}, \etc These help not only reduce the workload and burden of pending cases~\cite{njdg_v3} but also automate and standardise a number of legal processes. It is especially useful for countries like India, where documents are still unstructured and hand-written~\cite{paul-2022-pretraining}. While such tools warrant careful use, current evidence suggests they have limited direct impact on the justice system itself. On the other hand, more recently, there has been a rise in using AI for aiding judgment decisions~\cite{chakraborty2025well,westermann2024analyzing}. We recognize bail prediction as a highly sensitive application that requires continuous oversight (through lawmakers and independent technical audits) as well as clear regulatory frameworks before any real-world deployment. Our anticipatory audit represents, to our knowledge, the first systematic evaluation of AI for bail prediction that incorporates multimodal inputs, extending beyond prior unimodal analyses~\cite{kapoor2022hldc} and, thereby, simulating more realistic aspects of courtroom use. Given the scarcity of multimodal datasets for this domain, we curate our own dataset by combining Illinois DOC face images~\cite{illinoisimages} with textual case facts from the HLDC dataset~\cite{kapoor2022hldc}. Our findings suggest that such models, in their current form, are not suitable for deployment in real-world settings. Specifically, we observe that models are more likely to deny bail in cases where experienced judges have granted it, underscoring the importance of human-in-the-loop frameworks not only in deployment but also in system design. These observations highlight the need for carefully designed regulations, independently verifiable oversight mechanisms, and a cautious approach to the development and use of AI in sensitive domains such as bail prediction. Having said that, we also firmly believe that these models with proper interventions in place can be used as very efficient and effective \textit{assistive tools} across courtrooms.

%% file: sec/7_end.tex
\section{Conclusion}

Through this investigation, we aimed to shed light on the potentials and limitations of VLMs in real-world legal settings. Our findings have implications not only for the technical development of legal AI systems but also for the ethical and policy frameworks surrounding their deployment. Through detailed intervention and evaluation setups we show that \textit{by applying the correct interventions, we can bring out better performance from a VLM in a legal judgment prediction context}. Our work emphasizes the necessity of precedents in legal judgment prediction where AI models do not see any other evidences besides the provided case facts, restating the importance of this method implemented by many other contemporary works. We note that current VLMs along the suitable interventions can at best act as assistive tools in the courtroom and the final human emotive-cognitive delivery of justice is indispensable. 

Though the inclusion of multiple modalities, in terms of image, audio, video, text, is very imminent and practical in the legal context, we should be extra-aware before deploying such multimodal models. By introducing new modalities, the AI models become harder to interpret and easier to propagate and multiply the existing errors. While our multitude of intervention methods show a thorough improvement of the model behavior we believe that more aggressive research is needed to develop stronger interventions in the future before actual real-world deployment.

%% file: main.bbl
\begin{thebibliography}{49}
\providecommand{\natexlab}[1]{#1}
\providecommand{\url}[1]{\texttt{#1}}
\expandafter\ifx\csname urlstyle\endcsname\relax
  \providecommand{\doi}[1]{doi: #1}\else
  \providecommand{\doi}{doi: \begingroup \urlstyle{rm}\Url}\fi

\bibitem[Achiam et~al.(2023)Achiam, Adler, Agarwal, Ahmad, Akkaya, Aleman, Almeida, Altenschmidt, Altman, Anadkat, et~al.]{achiam2023gpt}
Josh Achiam, Steven Adler, Sandhini Agarwal, Lama Ahmad, Ilge Akkaya, Florencia~Leoni Aleman, Diogo Almeida, Janko Altenschmidt, Sam Altman, Shyamal Anadkat, et~al.
\newblock Gpt-4 technical report.
\newblock \emph{arXiv preprint arXiv:2303.08774}, 2023.

\bibitem[Akter et~al.(2025)Akter, {\c{C}}ano, Weber, Dobler, and Habernal]{akter2025comprehensive}
Mousumi Akter, Erion {\c{C}}ano, Erik Weber, Dennis Dobler, and Ivan Habernal.
\newblock A comprehensive survey on legal summarization: Challenges and future directions.
\newblock \emph{arXiv preprint arXiv:2501.17830}, 2025.

\bibitem[Aletras et~al.(2016)Aletras, Tsarapatsanis, Preo{\c{t}}iuc-Pietro, and Lampos]{aletras2016predicting}
Nikolaos Aletras, Dimitrios Tsarapatsanis, Daniel Preo{\c{t}}iuc-Pietro, and Vasileios Lampos.
\newblock Predicting judicial decisions of the european court of human rights: A natural language processing perspective.
\newblock \emph{PeerJ computer science}, 2:\penalty0 e93, 2016.

\bibitem[Asai et~al.(2023)Asai, Wu, Wang, Sil, and Hajishirzi]{asai2023self}
Akari Asai, Zeqiu Wu, Yizhong Wang, Avirup Sil, and Hannaneh Hajishirzi.
\newblock Self-rag: Self-reflective retrieval augmented generation.
\newblock In \emph{NeurIPS 2023 workshop on instruction tuning and instruction following}, 2023.

\bibitem[Brown et~al.(2020)Brown, Mann, Ryder, Subbiah, Kaplan, Dhariwal, Neelakantan, Shyam, Sastry, Askell, et~al.]{brown2020language}
Tom Brown, Benjamin Mann, Nick Ryder, Melanie Subbiah, Jared~D Kaplan, Prafulla Dhariwal, Arvind Neelakantan, Pranav Shyam, Girish Sastry, Amanda Askell, et~al.
\newblock Language models are few-shot learners.
\newblock \emph{Advances in neural information processing systems}, 33:\penalty0 1877--1901, 2020.

\bibitem[Chakraborty et~al.(2025)Chakraborty, Harit, and Ghosh]{chakraborty2025well}
Sagar Chakraborty, Gaurav Harit, and Saptarshi Ghosh.
\newblock How well do mllms understand handwritten legal documents? a novel dataset for benchmarking.
\newblock \emph{International Journal on Document Analysis and Recognition (IJDAR)}, pages 1--17, 2025.

\bibitem[Chalkidis et~al.(2020)Chalkidis, Fergadiotis, Malakasiotis, Aletras, and Androutsopoulos]{chalkidis2020legal}
Ilias Chalkidis, Manos Fergadiotis, Prodromos Malakasiotis, Nikolaos Aletras, and Ion Androutsopoulos.
\newblock Legal-bert: The muppets straight out of law school.
\newblock \emph{arXiv preprint arXiv:2010.02559}, 2020.

\bibitem[Contini et~al.(2024)Contini, Minissale, and Bergman~Blix]{Contini2024AI_Legal}
F. Contini, A. Minissale, and S. Bergman~Blix.
\newblock Artificial intelligence and real decisions: predictive systems and generative ai vs. emotive-cognitive legal deliberations.
\newblock \emph{Frontiers in Sociology}, 2024.

\bibitem[Cui et~al.(2023)Cui, Li, Yan, Chen, and Yuan]{cui2023chatlaw}
Jiaxi Cui, Zongjian Li, Yang Yan, Bohua Chen, and Li Yuan.
\newblock Chatlaw: Open-source legal large language model with integrated external knowledge bases.
\newblock \emph{CoRR}, 2023.

\bibitem[Dubey et~al.(2024)Dubey, Jauhri, Pandey, Kadian, Al-Dahle, Letman, Mathur, Schelten, Yang, Fan, et~al.]{dubey2024llama}
Abhimanyu Dubey, Abhinav Jauhri, Abhinav Pandey, Abhishek Kadian, Ahmad Al-Dahle, Aiesha Letman, Akhil Mathur, Alan Schelten, Amy Yang, Angela Fan, et~al.
\newblock The llama 3 herd of models.
\newblock \emph{arXiv e-prints}, pages arXiv--2407, 2024.

\bibitem[Farber(2025)]{Farber2025ForensicAI}
S. Farber.
\newblock Ai as a decision support tool in forensic image analysis: A pilot study on integrating large language models into crime scene investigation workflows.
\newblock \emph{Journal of Forensic Sciences}, 70\penalty0 (3):\penalty0 932--943, 2025.

\bibitem[Fisher()]{illinoisimages}
David~J. Fisher.
\newblock Illinois doc labeled faces dataset.
\newblock Kaggle.

\bibitem[Fullerton~Joireman(2006)]{fullerton2006evolution}
Sandra Fullerton~Joireman.
\newblock The evolution of the common law: Legal development in kenya and india.
\newblock \emph{Commonwealth \& Comparative Politics}, 44\penalty0 (2):\penalty0 190--210, 2006.

\bibitem[Gala et~al.(2023)Gala, Chitale, AK, Gumma, Doddapaneni, Kumar, Nawale, Sujatha, Puduppully, Raghavan, et~al.]{gala2023indictrans2}
Jay Gala, Pranjal~A Chitale, Raghavan AK, Varun Gumma, Sumanth Doddapaneni, Aswanth Kumar, Janki Nawale, Anupama Sujatha, Ratish Puduppully, Vivek Raghavan, et~al.
\newblock Indictrans2: Towards high-quality and accessible machine translation models for all 22 scheduled indian languages.
\newblock \emph{arXiv preprint arXiv:2305.16307}, 2023.

\bibitem[Hartsock and Rasool(2024)]{hartsock2024vision}
Iryna Hartsock and Ghulam Rasool.
\newblock Vision-language models for medical report generation and visual question answering: A review.
\newblock \emph{Frontiers in artificial intelligence}, 7:\penalty0 1430984, 2024.

\bibitem[Jiang et~al.(2023)Jiang, Sablayrolles, Mensch, Bamford, Chaplot, de~las Casas, Bressand, Lengyel, Lample, Saulnier, Lavaud, Lachaux, Stock, Scao, Lavril, Wang, Lacroix, and Sayed]{jiang2023mistral7b}
Albert~Q. Jiang, Alexandre Sablayrolles, Arthur Mensch, Chris Bamford, Devendra~Singh Chaplot, Diego de~las Casas, Florian Bressand, Gianna Lengyel, Guillaume Lample, Lucile Saulnier, Lélio~Renard Lavaud, Marie-Anne Lachaux, Pierre Stock, Teven~Le Scao, Thibaut Lavril, Thomas Wang, Timothée Lacroix, and William~El Sayed.
\newblock Mistral 7b, 2023.

\bibitem[Jiang and Yang(2023)]{jiang2023legal}
Cong Jiang and Xiaolei Yang.
\newblock Legal syllogism prompting: Teaching large language models for legal judgment prediction.
\newblock In \emph{Proceedings of the nineteenth international conference on artificial intelligence and law}, pages 417--421, 2023.

\bibitem[Kapoor et~al.(2022)Kapoor, Dhawan, Goel, Arjun, Bhatnagar, Agrawal, Agrawal, Bhattacharya, Kumaraguru, and Modi]{kapoor2022hldc}
Arnav Kapoor, Mudit Dhawan, Anmol Goel, TH Arjun, Akshala Bhatnagar, Vibhu Agrawal, Amul Agrawal, Arnab Bhattacharya, Ponnurangam Kumaraguru, and Ashutosh Modi.
\newblock Hldc: Hindi legal documents corpus.
\newblock \emph{arXiv preprint arXiv:2204.00806}, 2022.

\bibitem[Katz et~al.(2017)Katz, Bommarito~II, and Blackman]{katz2017general}
Daniel~Martin Katz, Michael~J Bommarito~II, and Josh Blackman.
\newblock A general approach for predicting the behavior of the supreme court of the united states.
\newblock \emph{PloS one}, 12\penalty0 (4):\penalty0 e0174698, 2017.

\bibitem[Laurençon et~al.(2024)Laurençon, Marafioti, Sanh, and Tronchon]{laurençon2024building}
Hugo Laurençon, Andrés Marafioti, Victor Sanh, and Léo Tronchon.
\newblock Building and better understanding vision-language models: insights and future directions., 2024.

\bibitem[Liu et~al.(2023)Liu, Li, Wu, and Lee]{liu2023visual}
Haotian Liu, Chunyuan Li, Qingyang Wu, and Yong~Jae Lee.
\newblock Visual instruction tuning.
\newblock \emph{Advances in neural information processing systems}, 36:\penalty0 34892--34916, 2023.

\bibitem[Liu et~al.(2024)Liu, Li, Li, Li, Zhang, Shen, and Lee]{liu2024llavanext}
Haotian Liu, Chunyuan Li, Yuheng Li, Bo Li, Yuanhan Zhang, Sheng Shen, and Yong~Jae Lee.
\newblock Llavanext: Improved reasoning, ocr, and world knowledge, 2024.

\bibitem[Lyu et~al.(2023)Lyu, Hao, Wang, Zhao, Gao, Ren, Chen, Wang, and Ren]{lyu2023multi}
Yougang Lyu, Jitai Hao, Zihan Wang, Kai Zhao, Shen Gao, Pengjie Ren, Zhumin Chen, Fang Wang, and Zhaochun Ren.
\newblock Multi-defendant legal judgment prediction via hierarchical reasoning.
\newblock \emph{arXiv preprint arXiv:2312.05762}, 2023.

\bibitem[Mahapatra et~al.(2023)Mahapatra, Datta, Soni, Goswami, and Ghosh]{mahapatra2023milpac}
Sayan Mahapatra, Debtanu Datta, Shubham Soni, Adrijit Goswami, and Saptarshi Ghosh.
\newblock Milpac: A novel benchmark for evaluating translation of legal text to indian languages.
\newblock \emph{arXiv preprint arXiv:2310.09765}, 2023.

\bibitem[Malik et~al.(2021)Malik, Sanjay, Nigam, Ghosh, Guha, Bhattacharya, and Modi]{malik2021ildc}
Vijit Malik, Rishabh Sanjay, Shubham~Kumar Nigam, Kripa Ghosh, Shouvik~Kumar Guha, Arnab Bhattacharya, and Ashutosh Modi.
\newblock Ildc for cjpe: Indian legal documents corpus for court judgment prediction and explanation.
\newblock \emph{arXiv preprint arXiv:2105.13562}, 2021.

\bibitem[Morin-Martel(2024)]{morin2024machine}
Alexis Morin-Martel.
\newblock Machine learning in bail decisions and judges’ trustworthiness.
\newblock \emph{Ai \& Society}, 39\penalty0 (4):\penalty0 2033--2044, 2024.

\bibitem[Nigam et~al.(2024{\natexlab{a}})Nigam, Deroy, Maity, and Bhattacharya]{nigam2024rethinking}
Shubham~Kumar Nigam, Aniket Deroy, Subhankar Maity, and Arnab Bhattacharya.
\newblock Rethinking legal judgement prediction in a realistic scenario in the era of large language models.
\newblock \emph{arXiv preprint arXiv:2410.10542}, 2024{\natexlab{a}}.

\bibitem[Nigam et~al.(2024{\natexlab{b}})Nigam, Patnaik, Mishra, Shallum, Ghosh, and Bhattacharya]{nigam2024nyayaanumana}
Shubham~Kumar Nigam, Balaramamahanthi~Deepak Patnaik, Shivam Mishra, Noel Shallum, Kripabandhu Ghosh, and Arnab Bhattacharya.
\newblock Nyayaanumana \& inlegalllama: the largest indian legal judgment prediction dataset and specialized language model for enhanced decision analysis.
\newblock \emph{arXiv preprint arXiv:2412.08385}, 2024{\natexlab{b}}.

\bibitem[Nigam et~al.(2025)Nigam, Patnaik, Mishra, Thomas, Shallum, Ghosh, and Bhattacharya]{nigam2025nyayarag}
Shubham~Kumar Nigam, Balaramamahanthi~Deepak Patnaik, Shivam Mishra, Ajay~Varghese Thomas, Noel Shallum, Kripabandhu Ghosh, and Arnab Bhattacharya.
\newblock Nyayarag: Realistic legal judgment prediction with rag under the indian common law system.
\newblock \emph{arXiv preprint arXiv:2508.00709}, 2025.

\bibitem[of~India()]{njdg_v3}
District~Court of India.
\newblock National judicial data grid.

\bibitem[Paul et~al.(2023)Paul, Mandal, Goyal, and Ghosh]{paul-2022-pretraining}
Shounak Paul, Arpan Mandal, Pawan Goyal, and Saptarshi Ghosh.
\newblock Pre-trained language models for the legal domain: A case study on indian law.
\newblock In \emph{Proceedings of 19th International Conference on Artificial Intelligence and Law - ICAIL 2023}, 2023.

\bibitem[Paul et~al.(2024)Paul, Bhatt, Goyal, and Ghosh]{paul2024legal}
Shounak Paul, Rajas Bhatt, Pawan Goyal, and Saptarshi Ghosh.
\newblock Legal statute identification: A case study using state-of-the-art datasets and methods.
\newblock In \emph{Proceedings of the 47th International ACM SIGIR Conference on Research and Development in Information Retrieval}, pages 2231--2240, 2024.

\bibitem[Poudyal et~al.(2020)Poudyal, {\v{S}}avelka, Ieven, Moens, Goncalves, and Quaresma]{poudyal2020echr}
Prakash Poudyal, Jarom{\'\i}r {\v{S}}avelka, Aagje Ieven, Marie~Francine Moens, Teresa Goncalves, and Paulo Quaresma.
\newblock Echr: Legal corpus for argument mining.
\newblock In \emph{Proceedings of the 7th Workshop on Argument Mining}, pages 67--75, 2020.

\bibitem[Radford et~al.(2021)Radford, Kim, Hallacy, Ramesh, Goh, Agarwal, Sastry, Askell, Mishkin, Clark, et~al.]{radford2021learning}
Alec Radford, Jong~Wook Kim, Chris Hallacy, Aditya Ramesh, Gabriel Goh, Sandhini Agarwal, Girish Sastry, Amanda Askell, Pamela Mishkin, Jack Clark, et~al.
\newblock Learning transferable visual models from natural language supervision.
\newblock In \emph{International conference on machine learning}, pages 8748--8763. PmLR, 2021.

\bibitem[Ruggeri et~al.(2023)Ruggeri, Nozza, et~al.]{ruggeri2023multi}
Gabriele Ruggeri, Debora Nozza, et~al.
\newblock A multi-dimensional study on bias in vision-language models.
\newblock In \emph{Findings of the Association for Computational Linguistics: ACL 2023}. Association for Computational Linguistics, 2023.

\bibitem[Sansone and Sperl{\'\i}(2022)]{sansone2022legal}
Carlo Sansone and Giancarlo Sperl{\'\i}.
\newblock Legal information retrieval systems: state-of-the-art and open issues.
\newblock \emph{Information Systems}, 106:\penalty0 101967, 2022.

\bibitem[Shareghi et~al.(2024)Shareghi, Han, and Burgess]{shareghi2024methods}
Ehsan Shareghi, Jiuzhou Han, and Paul Burgess.
\newblock Methods for legal citation prediction in the age of llms: An australian law case study.
\newblock \emph{arXiv e-prints}, pages arXiv--2412, 2024.

\bibitem[Stamatakis et~al.(2025)Stamatakis, Berger, Wartena, Ewerth, and Hoppe]{stamatakis2025enhancing}
Markos Stamatakis, Joshua Berger, Christian Wartena, Ralph Ewerth, and Anett Hoppe.
\newblock Enhancing the learning experience: Using vision-language models to generate questions for educational videos.
\newblock In \emph{International Conference on Artificial Intelligence in Education}, pages 305--319. Springer, 2025.

\bibitem[Team(2025)]{qwen2.5-VL}
Qwen Team.
\newblock Qwen2.5-vl, 2025.

\bibitem[Touvron et~al.(2023)Touvron, Martin, Stone, Albert, Almahairi, Babaei, Bashlykov, Batra, Bhargava, Bhosale, et~al.]{touvron2023llama}
Hugo Touvron, Louis Martin, Kevin Stone, Peter Albert, Amjad Almahairi, Yasmine Babaei, Nikolay Bashlykov, Soumya Batra, Prajjwal Bhargava, Shruti Bhosale, et~al.
\newblock Llama 2: Open foundation and fine-tuned chat models.
\newblock \emph{arXiv preprint arXiv:2307.09288}, 2023.

\bibitem[Wang et~al.(2025)Wang, Gao, Gu, Pu, Cui, Wei, Liu, Jing, Ye, Shao, et~al.]{wang2025internvl3_5}
Weiyun Wang, Zhangwei Gao, Lixin Gu, Hengjun Pu, Long Cui, Xingguang Wei, Zhaoyang Liu, Linglin Jing, Shenglong Ye, Jie Shao, et~al.
\newblock Internvl3.5: Advancing open-source multimodal models in versatility, reasoning, and efficiency.
\newblock \emph{arXiv preprint arXiv:2508.18265}, 2025.

\bibitem[Wang et~al.(2024)Wang, Wang, Le, Zheng, Mishra, Perot, Zhang, Mattapalli, Taly, Shang, et~al.]{wang2024speculative}
Zilong Wang, Zifeng Wang, Long Le, Huaixiu~Steven Zheng, Swaroop Mishra, Vincent Perot, Yuwei Zhang, Anush Mattapalli, Ankur Taly, Jingbo Shang, et~al.
\newblock Speculative rag: Enhancing retrieval augmented generation through drafting.
\newblock \emph{arXiv preprint arXiv:2407.08223}, 2024.

\bibitem[Westermann and Savelka(2024)]{westermann2024analyzing}
Hannes Westermann and Jaromir Savelka.
\newblock Analyzing images of legal documents: Toward multi-modal llms for access to justice.
\newblock \emph{arXiv preprint arXiv:2412.15260}, 2024.

\bibitem[Wu et~al.(2023)Wu, Zhou, Liu, Lu, Liu, Zhang, Sun, Wu, and Kuang]{wu2023precedent}
Yiquan Wu, Siying Zhou, Yifei Liu, Weiming Lu, Xiaozhong Liu, Yating Zhang, Changlong Sun, Fei Wu, and Kun Kuang.
\newblock Precedent-enhanced legal judgment prediction with llm and domain-model collaboration.
\newblock \emph{arXiv preprint arXiv:2310.09241}, 2023.

\bibitem[Xiao et~al.(2018)Xiao, Zhong, Guo, Tu, Liu, Sun, Feng, Han, Hu, Wang, et~al.]{xiao2018cail2018}
Chaojun Xiao, Haoxi Zhong, Zhipeng Guo, Cunchao Tu, Zhiyuan Liu, Maosong Sun, Yansong Feng, Xianpei Han, Zhen Hu, Heng Wang, et~al.
\newblock Cail2018: A large-scale legal dataset for judgment prediction.
\newblock \emph{arXiv preprint arXiv:1807.02478}, 2018.

\bibitem[Xiao et~al.(2024)Xiao, Liu, Cheng, Yin, Liang, Li, Shao, Liu, and Tao]{xiao2024genderbias}
Yisong Xiao, Aishan Liu, QianJia Cheng, Zhenfei Yin, Siyuan Liang, Jiapeng Li, Jing Shao, Xianglong Liu, and Dacheng Tao.
\newblock Genderbias-$\backslash$emph $\{$VL$\}$: Benchmarking gender bias in vision language models via counterfactual probing.
\newblock \emph{arXiv preprint arXiv:2407.00600}, 2024.

\bibitem[Yuan et~al.(2019)Yuan, Wang, Fan, Bian, Yang, Wang, and Wang]{yuan2019automatic}
Lufeng Yuan, Jun Wang, Shifeng Fan, Yingying Bian, Binming Yang, Yueyue Wang, and Xiaobin Wang.
\newblock Automatic legal judgment prediction via large amounts of criminal cases.
\newblock In \emph{2019 IEEE 5th International Conference on Computer and Communications (ICCC)}, pages 2087--2091. IEEE, 2019.

\bibitem[Zafar et~al.(2017)Zafar, Valera, Gomez~Rodriguez, and Gummadi]{zafar:17}
Muhammad~Bilal Zafar, Isabel Valera, Manuel Gomez~Rodriguez, and Krishna~P. Gummadi.
\newblock Fairness beyond disparate treatment \& disparate impact: Learning classification without disparate mistreatment.
\newblock In \emph{Proceedings of the 26th International Conference on World Wide Web}, page 1171–1180, Republic and Canton of Geneva, CHE, 2017. International World Wide Web Conferences Steering Committee.

\bibitem[Zhao et~al.(2024)Zhao, Zhang, Yu, Wang, Geng, Fu, Yang, Zhang, Jiang, and Cui]{zhao2024retrieval}
Penghao Zhao, Hailin Zhang, Qinhan Yu, Zhengren Wang, Yunteng Geng, Fangcheng Fu, Ling Yang, Wentao Zhang, Jie Jiang, and Bin Cui.
\newblock Retrieval-augmented generation for ai-generated content: A survey.
\newblock \emph{arXiv preprint arXiv:2402.19473}, 2024.

\end{thebibliography}
